\documentclass[conference]{IEEEtran}
\usepackage{times}

\usepackage[numbers]{natbib}

\usepackage{amsmath}
\usepackage{amsfonts}
\usepackage{amssymb}
\usepackage{siunitx}
\usepackage{pifont}   % check mark
\usepackage{xfrac}

\usepackage{booktabs}
\usepackage{makecell}  % Line breaks in table cells
\usepackage[flushleft]{threeparttable}  % For notes below a table
\usepackage{multirow}
\usepackage{multicol}

\usepackage[dvipsnames]{xcolor}    % for colored text
\usepackage{colortbl}
\usepackage{textcomp}  % prevent warnings of gensymb package
\usepackage{gensymb}   % \degree
\usepackage{graphicx} % insert PDF as figure
\usepackage{flushend}

\usepackage[bookmarks=true]{hyperref}

\usepackage[capitalize]{cleveref}
\crefname{section}{Sec.}{Secs.}
\Crefname{section}{Section}{Sections}
\Crefname{table}{Table}{Tables}
\crefname{table}{Tab.}{Tabs.}

\DeclareMathOperator*{\argmax}{arg\,max}

\newcommand{\red}[1]{{\color{red} #1}}
\newcommand{\citem}[1]{\textbf{\red{[?]}}}

\newcommand{\grayrule}{\arrayrulecolor{black!30}\midrule\arrayrulecolor{black}}

\newcommand{\cmark}{\ding{51}}  % pick from here: https://i.stack.imgur.com/Vjm6r.png
\begin{document}

\title{LiDAR Registration with Visual Foundation Models}

% You will get a Paper-ID when submitting a pdf file to the conference system
% \author{Author Names Omitted for Anonymous Review. Paper-ID xx}

%\author{\authorblockN{Michael Shell}
%\authorblockA{School of Electrical and\\Computer Engineering\\
%Georgia Institute of Technology\\
%Atlanta, Georgia 30332--0250\\
%Email: mshell@ece.gatech.edu}
%\and
%\authorblockA{Twentieth Century Fox\\
%\and
%\authorblockN{James Kirk\\ and Montgomery Scott}
%\authorblockA{Starfleet Academy\\
%San Francisco, California 96678-2391\\

% avoiding spaces at the end of the author lines is not a problem with
% conference papers because we don't use \thanks or \IEEEmembership

%Homer Simpson\authorrefmark{2},
%Montgomery Scott\authorrefmark{3} and
%Eldon Tyrell\authorrefmark{4}}
%Telephone: (800) 555--1212, Fax: (888) 555--1212}
%\authorblockA{\authorrefmark{4}Tyrell Inc., 123 Replicant Street, Los Angeles, California 90210--4321}}

\author{\authorblockN{Niclas Vödisch\textsuperscript{\footnotesize 1,2},
Giovanni Cioffi\textsuperscript{\footnotesize 2},
Marco Cannici\textsuperscript{\footnotesize 2},
Wolfram Burgard\textsuperscript{\footnotesize 3}, and
Davide Scaramuzza\textsuperscript{\footnotesize 2}} \vspace{0.1cm}
\authorblockA{\textsuperscript{\footnotesize 1}University of Freiburg \quad\quad {\textsuperscript{\footnotesize 2}University of Zurich} \quad\quad {\textsuperscript{\footnotesize 3}University of Technology Nuremberg}}
}

%%%%%%%%%%%%%%%%%%%%%%%%%%%%%%%%%%%%%%%%%%%%%%%%%%%%%%%%%%%%%%%%%%%%%%%%%%%%%%%%

\maketitle

\begin{abstract}
LiDAR registration is a fundamental task in robotic mapping and localization. A critical component of aligning two point clouds is identifying robust point correspondences using point descriptors. This step becomes particularly challenging in scenarios involving domain shifts, seasonal changes, and variations in point cloud structures. These factors substantially impact both handcrafted and learning-based approaches.
In this paper, we address these problems by proposing to use DINOv2 features, obtained from surround-view images, as point descriptors. We demonstrate that coupling these descriptors with traditional registration algorithms, such as RANSAC or ICP, facilitates robust 6DoF alignment of LiDAR scans with 3D maps, even when the map was recorded more than a year before. Although conceptually straightforward, our method substantially outperforms more complex baseline techniques. In contrast to previous learning-based point descriptors, our method does not require domain-specific retraining and is agnostic to the point cloud structure, effectively handling both sparse LiDAR scans and dense 3D maps.
We show that leveraging the additional camera data enables our method to outperform the best baseline by $+24.8$ and $+17.3$ registration recall on the NCLT and Oxford RobotCar datasets.
We publicly release the registration benchmark and the code of our work on \mbox{\url{https://vfm-registration.cs.uni-freiburg.de}}.
% We will publicly release the registration benchmark and the code of our work upon acceptance of this manuscript (\textit{the code and a video is available for reviewers in the supplementary material}).
\looseness=-1

\end{abstract}

\IEEEpeerreviewmaketitle

\section{Introduction}
\label{sec:intro}

Aligning two point clouds to compute their relative 3D transformation is a critical task in numerous robotic applications, including LiDAR odometry~\cite{vizzo2023kissicp}, loop closure registration~\cite{arce2023padloc}, and map-based localization~\cite{hroob2024generalizable}.
In this work, we specifically discuss map-based localization, which not only generalizes the other aforementioned tasks but is also critical for improving the efficiency and autonomy of mobile robots in environments where pre-existing map data is available.

Although place recognition~\cite{keetha2024anyloc, kim2019scancontextimage} or GNSS readings can provide an approximate initial estimate, their accuracy is generally insufficient for obtaining precise 3D poses relative to the map. In contrast, global point cloud registration~\cite{fischler1987ransac, zhou2016fgr} enables accurate 3D localization but necessitates the identification of reliable point correspondences between the point clouds. 
These correspondences are typically established by iterating over all points in the source point cloud to identify the most similar counterparts in the target frame.
Similarity is assessed using point descriptors, which are abstract feature representations of a point, e.g., encoding the geometry of its local environment.
In scan-to-map registration, point descriptors must be as unique as possible since the number of potential combinations grows with $\mathcal{O}(m \cdot n)$, referring to the number of points in the scan and the map. An additional challenge arises from temporal changes in the environment, such as seasonal variations or ongoing construction, necessitating point descriptors that are robust to such changes for long-term applicability~\cite{bianco2016nclt}.

Given the significance of this task, numerous point descriptors have been proposed, encompassing both traditional handcrafted~\cite{rusu2009fpfh} and learning-based~\cite{choy2019fcgf, poiesi2021dip, zeng20173dmatch} designs, primarily relying on 3D geometric features.
While learning-based descriptors tend to exhibit greater expressiveness than handcrafted methods, they often fail to generalize effectively to out-of-training domains and different point cloud representations, such as between RGB-D data and LiDAR scans~\cite{poiesi2023gedi}.
\looseness=-1

\begin{figure}[t]
    \centering
    \includegraphics[width=\linewidth]{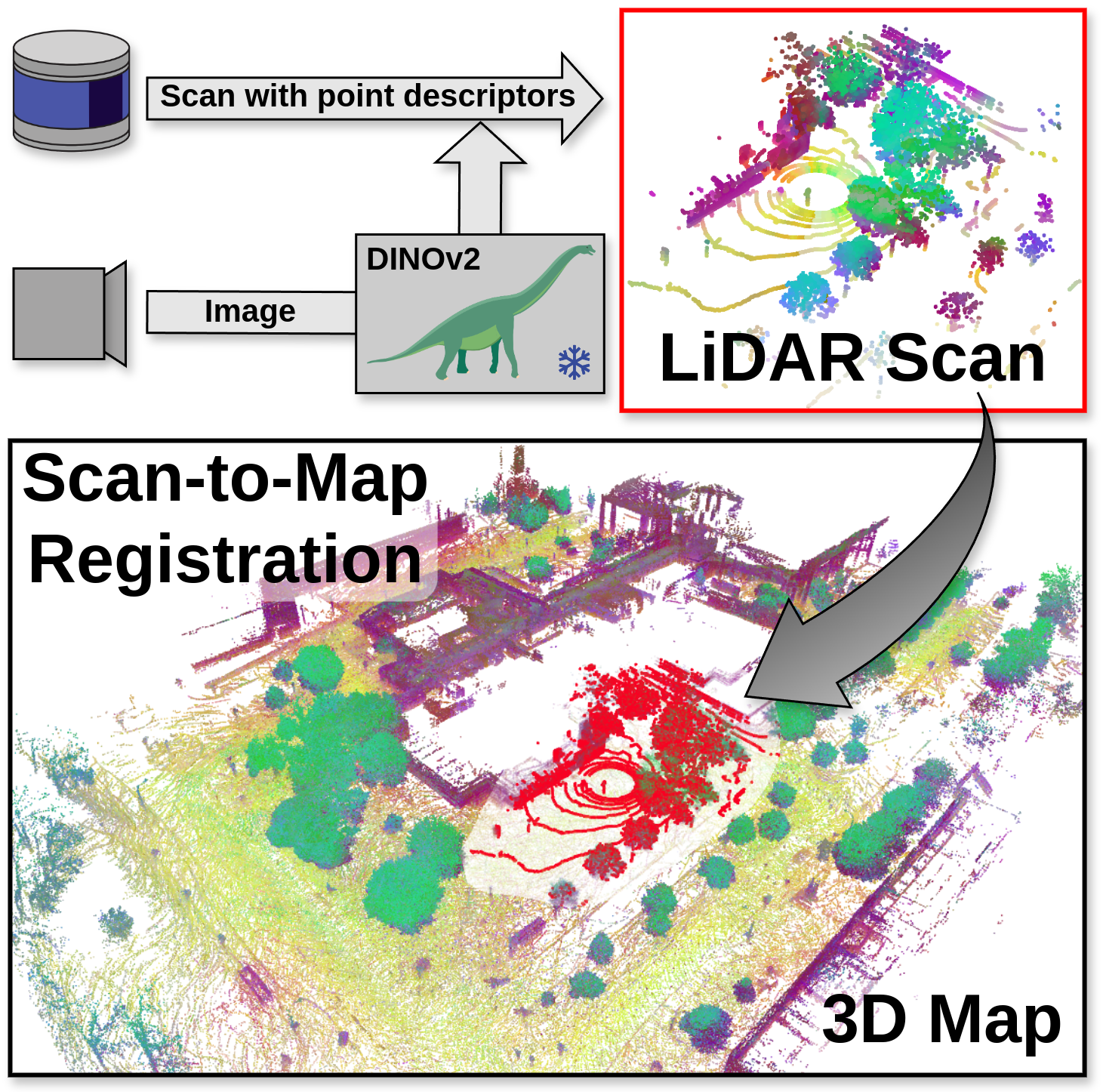}
    \vspace*{-.7cm}
    \caption{Initialization-free registration of a LiDAR scan to a large-scale 3D map requires highly expressive point descriptors. We demonstrate that DINOv2~\cite{oquab2024dinov2} features from surround-view images allow finding robust point correspondences, even with map data recorded more than a year before. In the map, the registered LiDAR scan is shown in red. The colors of the LiDAR scan (\textit{top right}) and the map (\textit{bottom}) are obtained using principal component analysis performed on the high-dimensional DINOv2 features.}
    \label{fig:teaser}
    \vspace*{-.4cm}
\end{figure}

In this work, we address the task of long-term scan-to-map registration by leveraging the advances made by recent visual foundation models. Our main contribution is to demonstrate that using DINOv2~\cite{oquab2024dinov2} features, obtained from surround-view images, as point descriptors allows finding highly robust point correspondences.
We argue that using the additional vision modality does not pose a large burden as this combination is a common sensor setup in mobile robotics~\cite{barnes2020robotcarradar, caesar2020nuscenes, bianco2016nclt} and the camera is relatively inexpensive compared to the LiDAR.

The key idea behind our approach is to leverage the superior generalization capabilities of recent visual foundation models, like DINOv2, compared to networks operating in the 3D space.
Furthermore, our approach is hence agnostic to the shape of a point cloud, enabling correspondence search between sparse LiDAR scans and dense 3D voxel-maps. Using DINOv2-based point descriptors effectively allows for implicit semantic matching between points and, importantly, does not require re-training an in-domain descriptor network.

We make three claims:
First, we demonstrate that coupling these descriptors with traditional registration algorithms, such as RANSAC~\cite{fischler1987ransac} or ICP~\cite{vizzo2023kissicp}, facilitates robust 3D localization in a map that was recorded over a year before~\cite{bianco2016nclt}.
Second, although conceptually simple, our method substantially outperforms more complex baseline techniques.
Third, our approach is robust to temporal changes in the environment that have occurred since the map was created.

We validate these claims through extensive experiments, showing that our approach outperforms the best baseline by $+24.8$ and $+17.3$ registration recall on the NCLT~\cite{bianco2016nclt} and Oxford Radar RobotCar~\cite{barnes2020robotcarradar} datasets.
To facilitate reproducibility and future research on long-term map registration, upon acceptance we will release our code along with instructions to re-create the evaluation scenes from our experiments.
To the best of our knowledge, this work presents the first approach to combining visual foundation models with traditional LiDAR registration techniques.
\looseness=-1

\section{Related Work}

Point cloud registration has been extensively studied by the research community across a diverse range of applications. Previous works have addressed alignment of relatively small 3D objects~\cite{breitenreicher2010robust}, mid-size indoor scenes~\cite{poiesi2021dip, zeng20173dmatch}, LiDAR odometry~\cite{cui2024sageicp, vizzo2023kissicp}, and scan registration to 3D maps~\cite{hroob2024generalizable}. Particularly the latter introduces further challenges, such as geometric and semantic discrepancies between the source and the target point clouds, arising from temporal changes in the environment since the creation of the reference map. While the majority of studies focus on object alignment or LiDAR odometry, only a few works~\cite{hroob2024generalizable, kim2019scancontextimage} explicitly target long-term scenarios. Typically, the point cloud registration problem is addressed in two stages: first, identifying point-to-point correspondences using point descriptors and, second, determining the six degrees of freedom (6DoF) transformation required to align the two point clouds. In the following paragraphs, we provide an overview of each of these steps. \looseness=-1

%%%%%%%%%%%%%%%%%%%%%%%%%%%%%%%%%%%%%%%%%%%%%%%%%%%%%%%%%%%%%%%%%%%%%%%%%%%%%%%%

\begin{figure*}[t]
    \centering
    \includegraphics[width=\linewidth]{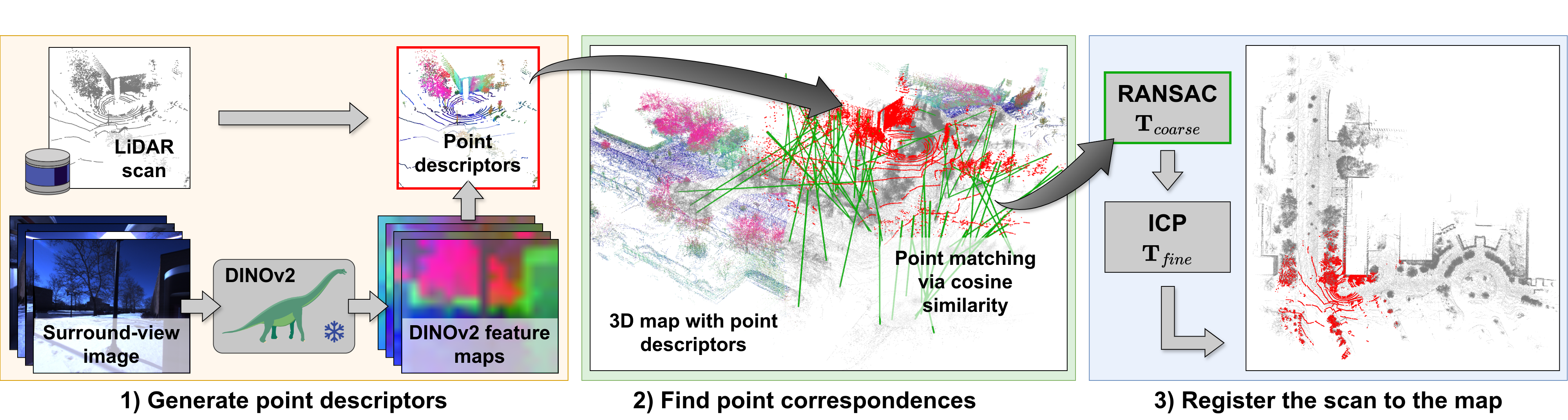}
    \vspace*{-.6cm}
    \caption{Overview of our proposed approach for 6DoF point cloud registration. First, we extract DINOv2~\cite{oquab2024dinov2} features from surround-view image data. These features are then attached to the point cloud as point descriptors via point-to-pixel projection. Second, we perform a point-wise similarity search using cosine similarity between the descriptors of the LiDAR scan and the descriptors of the voxelized 3D map. Finally, we use a traditional coarse-to-fine registration scheme with RANSAC~\cite{fischler1987ransac} and point-to-point ICP~\cite{vizzo2023kissicp} for obtaining a highly accurate pose estimate within the provided map frame.}
    \label{fig:overview}
    \vspace*{-.5cm}
\end{figure*}

%%%%%%%%%%%%%%%%%%%%%%%%%%%%%%%%%%%%%%%%%%%%%%%%%%%%%%%%%%%%%%%%%%%%%%%%%%%%%%%%

{\parskip=3pt
\noindent\textit{Point Descriptors:}
Point descriptors refer to an abstract representation of a 3D point that can be used to search for point-to-point correspondences between two point clouds. In contrast to a naive nearest-neighbor search in the Euclidean space, e.g., performed by the iterative closest point (ICP) algorithm~\cite{besl1992icp}, point descriptors allow finding global correspondences.
A classical, yet still commonly used~\cite{bai2021pointdsc, yang2021teaser} descriptor is the FPFH~\cite{rusu2009fpfh} descriptor that captures the geometry around a point by computing local feature histograms based on the angles to its neighboring points. FPFH is designed to provide both translational and rotational invariance.
In more recent years, many learning-based approaches have been proposed that employ deep neural networks for extracting point descriptors. These methods can generally be categorized into patch-based networks and fully convolutional networks.
3DMatch~\cite{zeng20173dmatch} pioneered the first category by learning local geometric patterns from volumetric patches around a point. While 3DMatch computes truncated distance function values from the patch, 3DSmoothNet~\cite{gojcic2019perfectmatch} uses a smoothed density value representation to achieve rotation invariance. Both DIP~\cite{poiesi2021dip} and GeDi~\cite{poiesi2023gedi} propose to follow a Siamese approach to train two neural networks with shared parameters and a contrastive loss on the patches. 
Unlike patch-based approaches, fully convolutional networks employ such a contrastive loss directly on the point level as first proposed by FCGF~\cite{choy2019fcgf}, which is commonly used by many registration algorithms~\cite{bai2021pointdsc}. While early convolutional descriptors were either computed for all points of a point cloud or a randomly sampled subset, later works such as D3Feat~\cite{bai2020d3feat} include keypoint detection schemes. A particular challenge of operating on individual points instead of patches is to achieve density invariance. Therefore, GCL~\cite{Liu2023gcl} focuses on low-overlap scenarios, e.g., required for early loop closure registration in LiDAR SLAM.
Although learning-based descriptors have shown impressive performance, a substantial drawback is their poor generalization between different training and testing domains, e.g., \mbox{RGB-D} data versus LiDAR scans or aligning individual objects versus large-scale outdoor scenes.

The majority of point descriptors focus on encoding only geometric information neglecting semantic hints. Especially in the context of autonomous driving, a few works have proposed to include information from semantic segmentation. For instance, SAGE-ICP~\cite{cui2024sageicp} extends the point correspondence search of ICP with a hard rejection scheme if the candidate points belong to different semantic classes. The transformer-based PADLoC~\cite{arce2023padloc} exploits panoptic information during the training phase to avoid wrong matches. Nonetheless, a major barrier to including semantic information is the lack of 3D segmentation networks that generalize well across domains requiring cost-intense retraining. In this work, we exploit the recent advances in the vision domain by proposing to extract point descriptors using a visual foundation model~\cite{oquab2024dinov2}. In contrast to the hard matching of discrete semantic classes, our method enables searching more soft correspondences while considering the scene semantics.
}

%%%%%%%%%%%%%%%%%%%%%%%%%%%%%%%%%%%%%%%%%%%%%%%%%%%%%%%%%%%%%%%%%%%%%%%%%%%%%%%%

{\parskip=3pt
\noindent\textit{Point Cloud Registration:}
Algorithms for point cloud registration can be categorized into local and global registration schemes. Whereas local methods require an accurate initial guess, global registration often assumes given point correspondences based on the aforementioned point descriptors. Often, both types are combined in a coarse-to-fine manner to achieve global registration with the high performance of local approaches such as ICP~\cite{besl1992icp} or NDT~\cite{biber2003ndt}.
To obtain a sufficient coarse registration, a main requirement for global registration schemes is outlier rejection. The most popular traditional method for this task is still RANSAC~\cite{fischler1987ransac}, including its more recent variants~\cite{barath2022magsac}. However, the major drawbacks of RANSAC are slow convergence and low accuracy in the presence of large outlier rates, which are commonly faced in point cloud registration. Fast global registration (FGR)~\cite{zhou2016fgr} aims to overcome these problems by optimizing a robust objective function that is defined densely over the surfaces. TEASER~\cite{yang2021teaser} proposes a certifiable algorithm that decouples scale, rotation, and translation estimation.
Similar to other fields, recent outlier rejection schemes attempt to improve their performance via deep learning. Both deep global registration (DGR)~\cite{choy2020dgr} and 3DRegNet~\cite{pais20203dregnet} formulate outlier rejection as a point-based classification problem. PointDSC~\cite{bai2021pointdsc} extends this idea by including the spatial consistency between inlier correspondences when applying rigid Euclidean transformations.
In this work, we demonstrate that coupling our proposed point descriptors with traditional registration algorithms, such as RANSAC or ICP, enables robust point cloud registration.
}

\section{Technical Approach}

In this section, we first formally define the problem addressed in this work. Then, we explain how to extract the point descriptors based on a visual foundation model. Finally, we elaborate on how we employ these descriptors for robust scan-to-map registration. We illustrate the separate steps in \cref{fig:overview}. \looseness=-1

%%%%%%%%%%%%%%%%%%%%%%%%%%%%%%%%%%%%%%%%%%%%%%%%%%%%%%%%%%%%%%%%%%%%%%%%%%%%%%%%

\subsection{Problem Definition}
\label{ssec:problem-definition}

In this work, we consider the following scenario:
A voxelized 3D map $M \in \mathbb{R}^{n \times 3}$ is provided, where $n$ denotes the number of 3D points stored in the map. At test-time, we receive a LiDAR scan $S \in \mathbb{R}^{m \times 3}$ composed of $m$ 3D points. Furthermore, for both $M$ and $S$, corresponding surround-view RGB camera data is available. The goal is to find the six degrees of freedom (6DoF) transform $\mathbf{T} \in \text{SE}(3)$ that correctly registers the LiDAR scan to the map.
We further assume a rough initial position $\hat{P} \in \mathbb{R}^3$ is given within approximately \SI{100}{\meter} around the true position, reducing the size of the relevant part of the map to $k << m$ while preserving $k >> n$. Such an initial position could be obtained via place recognition~\cite{keetha2024anyloc, kim2022scancontextpp} or GNSS readings.
Importantly, in long-term scan-to-map registration, the LiDAR scan can be recorded a considerable amount of time after the map was created, i.e., there might be a semantic and geometric discrepancy between the 3D map representation and the current state of the environment.

%%%%%%%%%%%%%%%%%%%%%%%%%%%%%%%%%%%%%%%%%%%%%%%%%%%%%%%%%%%%%%%%%%%%%%%%%%%%%%%%

\subsection{Point Descriptor Extraction}
\label{ssec:point-descriptors}

In this paragraph, we describe step 1) of \cref{fig:overview}.
While we extract the point descriptors of the LiDAR scan~$S$ at test-time, we pre-compute the descriptors of the map $M$ in an offline fashion.
Commonly, 3D point-based mapping approaches rely on the concept of keyframes to frequently identify LiDAR scans that are eventually accumulated into a single voxelized point cloud, i.e., a map is formally composed of individual LiDAR scans $M = \bigcup_j M_j$. 
In the following, we hence use the general notation of a point cloud $C \in \mathbb{R}^{l \times 3}$ to refer to either $S$ or $M_j$. Each point cloud can be associated with a surround-view RGB image taken at the same time as $C$. We denote this image as $I \in \mathbb{N}^{h \times w \times 3}$, where $h$ and $w$ represent its height and width, respectively.
First, we feed $I$ through a frozen DINOv2~\cite{oquab2024dinov2} model to generate a dense 2D feature map $F \in \mathbb{R}^{h \times w \times d}$, e.g., $d = 384$ for the model type ViT-S/14. The core idea of using DINOv2 is to capture the semantics of the scene without an explicit assignment to discrete semantic classes~\cite{cui2024sageicp} while leveraging its generalization capabilities across cameras, weather, and illumination conditions~\cite{keetha2024anyloc}. Second, we employ point-to-pixel projection via known extrinsic calibration parameters to convert each point $p \in C$ into pixel coordinates of $I$. Finally, we use the DINOv2 feature of the respective RGB pixel as the descriptor $\text{desc}(\cdot)$ of point $p$. Formally, \looseness=-1
\begin{align}
    F &= \texttt{DINOv2}(I) \, , \\
    \text{desc}(p) &= F \left[ \Pi(p) \right] \, ,
\end{align}
where $\Pi(\cdot): \mathbb{R}^3 \to \mathbb{N}^2$ is the point-to-pixel projection function and $[\cdot]: \mathbb{N}^2 \to \mathbb{R}^d$ denotes the operator to access the DINOv2 feature of a given pixel. Consequently, we apply this step to all points in $C$ to obtain $C^D$:
\begin{equation}
    C^D = \{ \text{desc}(p) \, | \, \forall p \in C \}
\end{equation}
We hence retrieve the descriptors $M^D \in \mathbb{R}^{k \times d}$ corresponding to the map $M$ and, at test-time, the descriptors $S^D \in \mathbb{R}^{m \times d}$ for the current scan $S$.

%%%%%%%%%%%%%%%%%%%%%%%%%%%%%%%%%%%%%%%%%%%%%%%%%%%%%%%%%%%%%%%%%%%%%%%%%%%%%%%%

\subsection{Scan-to-Map Registration}
\label{ssec:registration}

As defined in \cref{ssec:problem-definition}, the goal of scan-to-map registration is to find the 6DoF transform that correctly represents the robot pose with respect to the coordinate system of the map. 
In this paragraph, we describe the corresponding steps 2) and 3) of \cref{fig:overview}.
First, we substantially downsample the LiDAR scan $S$ with point descriptors $S^D$ resulting in $\tilde{S} \in \mathbb{R}^{v \times 3}$ and $\tilde{S}^D \in \mathbb{R}^{v \times d}$ with $v \approx \sfrac{m}{80}$ to reduce the complexity of the subsequent steps.
Second, we search for point correspondences between the scan and the map using an efficient similarity search~\cite{douze2024faiss}.
For every point $p_s \in \tilde{S}$, we search for the point $p_m \in M$ that achieves the highest cosine similarity between the descriptors of both points. \looseness=-1
\begin{align}
    p_m 
    &= \argmax_{p \in M} \text{sim}_{\cos} \left( \text{desc}(p_s), \text{desc}(p) \right) \\
    &= \argmax_{p \in M} \frac{\text{desc}(p_s) \cdot \text{desc}(p)}{||\text{desc}(p_s)|| \, ||\text{desc}(p)||}
\end{align}
Finally, if the cosine similarity is greater than a threshold $\theta_{\cos} = 0.8$, we consider the pair $(p_s, p_m)$ a valid point correspondence.

To achieve global registration within the map, we run \mbox{3-point} RANSAC~\cite{fischler1987ransac} on the set of all valid point correspondences, resulting in a coarse initial transform $\mathbf{T}_\textit{coarse}$. For further refinement and accurate 6DoF registration, we employ classical point-to-point ICP~\cite{vizzo2023kissicp} based on the 3D points of the original LiDAR scan~$S$. That is, the DINOv2-based point descriptors are not used in this step. With ICP, we obtain the final 6DoF transform $\mathbf{T}_\textit{fine}$ that aligns the scan~$S$ with the map~$M$.
\looseness=-1

\section{Experiments}

The main focus of this work is to enable LiDAR scan-to-map registration that is robust to the challenges arising in long-term scenarios. In our experiments, we demonstrate the effectiveness and capabilities of our approach to support our key claims:
First, DINOv2~\cite{oquab2024dinov2} features can serve as point descriptors that can be integrated into traditional registration algorithms to facilitate robust 3D registration in a map that was recorded more than a year before.
Second, such a relatively straightforward approach outperforms more complex baseline techniques.
Third, these descriptors are robust to temporal changes in the environment that have occurred since the map was created.
We begin this section by describing the details of our experimental setup and defining the evaluation metrics used. Afterward, we provide results that support our claims and showcase the performance of our approach.

%%%%%%%%%%%%%%%%%%%%%%%%%%%%%%%%%%%%%%%%%%%%%%%%%%%%%%%%%%%%%%%%%%%%%%%%%%%%%%%%

\subsection{Experimental Setup}
\label{ssec:experimental-setup}

To verify our claims, we generate several scenes that present the problem formally defined in \cref{ssec:problem-definition}.
In particular, we identify the NCLT~\cite{bianco2016nclt} and the Oxford Radar RobotCar~\cite{barnes2020robotcarradar} as two of the few long-term datasets containing both LiDAR and surround-view RGB images. Note that neither is part of the LVD-142M dataset~\cite{oquab2024dinov2} used to train DINOv2. The NCLT dataset comprises a total of 27 sessions recorded on a university campus spread out over 15 months, i.e., covering various seasons, illuminations, and environmental conditions. The Oxford Radar RobotCar dataset comprises recordings from 7 different days spread out over 1.5 weeks. It captures vehicle-centric urban data with weather conditions ranging from sunny to varying degrees of overcast.
Both datasets contain global pose data that is consistent across recordings.
For each dataset, we select the first of the provided recordings for mapping, while other recordings are used for registration. As shown in \cref{tab:dataset-recordings}, we carefully select these recordings; first, to maximize spread over the dataset to increase the probability of semantic and geometric changes and, second, to cover all available seasonal and weather conditions.

To construct a scene, we sample a position $\rho_i$ from the mapping route. If all registration recordings contain a point cloud associated with a pose in the vicinity of $\rho_i$, we use these point clouds to create  $\{ \mathcal{S}_1, \dots, \mathcal{S}_r \}_i$ and $\{ \mathcal{S}^D_1, \dots, \mathcal{S}^D_r \}_i$, i.e., a set of LiDAR scans with corresponding DINOv2-based~\cite{oquab2024dinov2} point descriptors. As listed in \cref{tab:dataset-recordings}, we use $r=5$ for NCLT and $r=3$ for RobotCar.
For mapping, we select the point clouds along the route within a \SI{150}{\meter} radius around $\rho_i$ and with a distance of \SI{2}{\meter} between scans, simulating keyframes in LiDAR SLAM, and extract the point descriptors. That is, the maximum size of the map is $\SI{300}{\meter} \times \SI{300}{\meter}$. Since the accuracy of the global poses provided in the datasets is insufficient for point cloud accumulation, we refine them with KISS-ICP~\cite{vizzo2023kissicp}. Finally, we downsample the accumulated resulting point cloud to a voxel size of \SI{0.25}{\meter}, representing the 3D map $M_i$ with point descriptors $M^D_i$. Note that we do not remove potentially dynamic objects, e.g., cars, from the map.
To measure the registration error, we generate ground truth transformations $\{\mathbf{T}_1, \dots, \mathbf{T}_r\}$ by running point-to-point ICP initialized with the pose from the dataset and manually verifying the correct registration.
For both datasets, we construct 25 scenes, i.e., $i \in [1, 25]$, resulting in a sample size of 125 and 75 for NCLT and Oxford Radar RobotCar, respectively. For examples of the scenes, we refer to the qualitative results in \cref{fig:results-nclt,fig:results-robotcar}.
To facilitate the utilization of these scenes as a benchmark in future research, we provide a recreation script along with comprehensive instructions in our code release.

\begin{table}
\footnotesize
\centering
\caption{Recordings for Scene Generation}
\vspace{-0.2cm}
\label{tab:dataset-recordings}
\setlength\tabcolsep{3.0pt}
\begin{threeparttable}
    \begin{tabular}{ l | cc | cccc }
    \toprule
    \textbf{Recording date} & \textbf{Map.} & \textbf{Reg.} & Time & Sky & Foliage & Snow \\
    \midrule
    \multicolumn{3}{l}{\textbf{NCLT}} \\
    [.5ex]
    2012-01-08 & \cmark & & Midday & Partly cloudy & -- & -- \\
    2012-02-12 & & \cmark & Midday & Sunny & -- & \cmark \\
    2012-03-17 & & \cmark & Morning & Sunny & -- & -- \\
    2012-05-26 & & \cmark & Evening & Sunny & \cmark & -- \\
    2012-10-28 & & \cmark & Midday & Cloudy & -- & -- \\
    2013-04-05 & & \cmark & Afternoon & Sunny & -- & \cmark \\
    \midrule
    \multicolumn{4}{l}{\textbf{Oxford Radar RobotCar}} \\
    [.5ex]
    2019-01-10 & \cmark & & 11:46 & Cloudy & -- & -- \\
    2019-01-15 & & \cmark & 13:06 & Sunny & -- & -- \\
    2019-01-17 & & \cmark & 14:03 & Partly cloudy & -- & -- \\
    2019-01-18 & & \cmark & 15:20 & Cloudy & -- & -- \\
    \bottomrule
    \end{tabular}
    Overview of the recordings from the NCLT~\cite{bianco2016nclt} and the Oxford Radar RobotCar~\cite{barnes2020robotcarradar} datasets used for mapping (\textit{Map.}) and scan registration (\textit{Reg.}).
\end{threeparttable}
\end{table}

%%%%%%%%%%%%%%%%%%%%%%%%%%%%%%%%%%%%%%%%%%%%%%%%%%%%%%%%%%%%%%%%%%%%%%%%%%%%%%%%

\begin{table*}[t]
\footnotesize
\centering
\caption{Scan-to-Map Registration}
\vspace{-0.2cm}
\label{tab:scan-to-map-exp}
\setlength\tabcolsep{5.0pt}
\begin{threeparttable}
    \begin{tabular}{ ll | cccc | cccc }
        \toprule
        \multicolumn{2}{l|}{\textbf{Method}} & \multicolumn{4}{c|}{\textbf{NCLT}} & \multicolumn{4}{c}{\textbf{Oxford Radar RobotCar}} \\
        Registration & Descriptor & RTE [m] & RRE [$\degree$] & RR [\%] & ICP-RR [\%] & RTE [m] & RRE [$\degree$] & RR [\%] & ICP-RR [\%] \\
        \midrule
        RANSAC & FPFH~\cite{rusu2009fpfh} & 47.51$\pm$46.54 & 59.95$\pm$62.21 & 0.00 & 37.60 & 32.56$\pm$40.18 & 36.87$\pm$57.25 & 0.00 & 40.00 \\
        TEASER++~\cite{yang2021teaser} & FPFH~\cite{rusu2009fpfh} & 60.15$\pm$34.94 & 121.67$\pm$48.95 & 0.00 & 0.00 & 69.51$\pm$41.98 & 129.01$\pm$49.91 & 0.00 & 0.00 \\
        PointDSC~\cite{bai2021pointdsc} & FPFH~\cite{rusu2009fpfh} & 103.16$\pm$48.70 & 129.75$\pm$47.90 & 0.00 & 0.00 & 65.28$\pm$45.19 & 128.01$\pm$52.22 & 1.33 & 2.67 \\
        RANSAC & DIP~\cite{poiesi2021dip} & 42.94$\pm$45.26 & 49.86$\pm$56.08 & 0.00 & 40.00 & 24.52$\pm$32.96 & 45.08$\pm$63.06 & 0.00 & 40.00 \\
        RANSAC & GeDi~\cite{poiesi2023gedi} & 48.93$\pm$45.12 & 66.75$\pm$65.59 & 0.00 & 32.00 & 23.50$\pm$31.15 & 41.26$\pm$62.85 & 0.00 & 44.00 \\
        RANSAC & FCGF~\cite{choy2019fcgf} & 22.74$\pm$33.53 & 31.20$\pm$48.44 & 0.80 & 64.00 & 20.34$\pm$32.29 & 37.92$\pm$62.81 & 1.33 & 49.33 \\
        PointDSC~\cite{bai2021pointdsc} & FCGF~\cite{choy2019fcgf} & 69.50$\pm$48.40 & 96.86$\pm$51.55 & 0.00 & 12.00 & 73.58$\pm$45.43 & 93.51$\pm$64.32 & 0.00 & 0.00 \\
        RANSAC & SpinNet~\cite{ao2021spinnet} & 33.70$\pm$42.47 & 41.00$\pm$55.14 & 0.00 & 54.40 & 23.39$\pm$35.86 & 30.57$\pm$56.48 & 0.00 & 56.00 \\
        RANSAC & GCL~\cite{Liu2023gcl} & 15.35$\pm$26.84 & 23.67$\pm$45.83 & 13.60 & 75.20 & 11.81$\pm$25.09 & 21.84$\pm$48.07 & 30.67 & 77.33 \\
        \grayrule
        \multicolumn{2}{l}{\textit{Our method}:} \\
        [.25ex]
        TEASER++~\cite{yang2021teaser} & DINOv2~\cite{oquab2024dinov2} & 5.29$\pm$10.02 & 27.98$\pm$57.09 & 18.40 & 83.20 & 13.45$\pm$34.93 & 22.18$\pm$53.69 & 49.33 & 81.33 \\
        RANSAC & DINOv2~\cite{oquab2024dinov2} & 0.40$\pm$0.31 & 1.01$\pm$0.84 & 77.60 & \textbf{100.00} & 3.36$\pm$13.65 & 5.20$\pm$26.25 & 82.67 & \textbf{94.67} \\
        RANSAC + ICP & DINOv2~\cite{oquab2024dinov2} & \textbf{0.01$\pm$0.02} & \textbf{0.03$\pm$0.06} & \textbf{100.00} & \textbf{100.00} & \textbf{3.12$\pm$13.63} & \textbf{4.52$\pm$26.21} & \textbf{94.67} & \textbf{94.67} \\
        \bottomrule
    \end{tabular}
    \footnotesize
    We report the mean and standard deviation of the relative translation error~(RTE) and the relative rotation error~(RRE) as defined in \cref{ssec:metrics}. The registration recall~(RR) denotes the success rate, where success is defined as $\text{RTE} < \SI{0.6}{\meter}$ and $\text{RRE} < \SI{1.5}{\degree}$. The recall after refinement with ICP is listed as ICP-RR.
    DIP and GeDi are trained on the 3DMatch dataset~\cite{zeng20173dmatch} (RGB-D data). FCGF, SpinNet, GCL, and PointDSC are trained on the KITTI dataset~\cite{geiger2012kitti} (LiDAR scans). \looseness=-1
\end{threeparttable}
\vspace*{-.3cm}
\end{table*}

\subsection{Evaluation Metrics}
\label{ssec:metrics}

Similar to prior work~\cite{ao2021spinnet, bai2020d3feat, bai2021pointdsc, choy2019fcgf}, we use the following evaluation metrics:
First, the relative translation error~(RTE) measured in meters.
Second, the relative rotation error~(RRE) measured in degrees.
Formally, the RTE and RRE are defined as:\looseness=-1
\begin{align}
    \text{RTE} &= ||\mathbf{\hat{t}} - \mathbf{t}||_2 \\
    \text{RRE} &= \arccos \left( \frac{\text{tr}(\mathbf{\hat{R}}^\intercal \mathbf{R}) - 1}{2} \right) \, ,
\end{align}
where the transform $\mathbf{T} \in \text{SE}(3)$ is decomposed into a translation $\mathbf{t}$ and rotation $\mathbf{R}$. The hat ($\hat{\cdot}$) denotes the estimated transform. The operators $\text{tr}(\cdot)$ and $(\cdot)^\intercal$ are the trace and transpose of a matrix.
Third, we report the registration recall~(RR) denoting the percentage of successful registrations, i.e., both the RTE and RRE are below a given threshold. While previous works have used different thresholds~\cite{ao2021spinnet, bai2020d3feat, bai2021pointdsc, Liu2023gcl}, we adopt the criterion of Liu~\textit{et~al.}~\cite{Liu2023gcl} as it meets the expected error range of many baseline techniques in the more simple scan-to-scan registration tasks (see \cref{ssec:exp-scan-to-map}). That is, a registration is considered a success if $\text{RTE} < \SI{0.6}{\meter}$ and $\text{RRE} < \SI{1.5}{\degree}$.
Finally, we investigate whether the accuracy of the descriptor-based global registration is sufficient to initialize ICP. We measure its performance by recomputing the registration recall after ICP-based pose refinement~(ICP-RR).

%%%%%%%%%%%%%%%%%%%%%%%%%%%%%%%%%%%%%%%%%%%%%%%%%%%%%%%%%%%%%%%%%%%%%%%%%%%%%%%%

\subsection{Scan-to-Map Registration}
\label{ssec:exp-scan-to-map}

We compare our proposed approach to a variety of baselines that can be categorized as follows:
(1) the popular handcrafted descriptor FPFH~\cite{rusu2009fpfh} coupled with three outlier rejection schemes, namely, RANSAC~\cite{fischler1987ransac}, TEASER++~\cite{yang2021teaser}, and PointDSC~\cite{bai2021pointdsc}, which is a learning-based approach trained on the KITTI dataset~\cite{geiger2012kitti};
(2) the learning-based descriptors DIP~\cite{poiesi2021dip} and GeDi~\cite{poiesi2023gedi}, which are trained on RGB-D data of the 3DMatch dataset~\cite{zeng20173dmatch} but claimed generalization to LiDAR scans~\cite{poiesi2021dip, poiesi2023gedi};
(3) the learning-based descriptors FCGF~\cite{choy2019fcgf}, SpinNet~\cite{ao2021spinnet}, and GCL~\cite{Liu2023gcl}, which are trained on the KITTI dataset~\cite{geiger2012kitti}. For categories (2) and (3), we predominantly rely on the RANSAC algorithm for point cloud registration and follow prior work~\cite{bai2020d3feat, zeng20173dmatch} using 50,000 iterations without early stopping. For the baselines, we use the top 5,000 point correspondences. For SpinNet~\cite{ao2021spinnet}, we compute the descriptors only for 7,500 randomly sampled points due to GPU memory constraints~(\SI{16}{\giga\byte}).

\begin{table}
\footnotesize
\centering
\caption{Scan-to-Scan Registration}
\vspace{-0.2cm}
\label{tab:sanity-exp}
\setlength\tabcolsep{3.0pt}
\begin{threeparttable}
    \begin{tabular}{ ll | cc | cc }
        \toprule
        \multicolumn{2}{l|}{\textbf{Method}} & \multicolumn{2}{c|}{\textbf{KITTI}} & \multicolumn{2}{c}{\textbf{NCLT}} \\
        Registration & Descriptor & RTE [m] & RRE [°] & RTE [m] & RRE [°] \\
        \midrule
        \multicolumn{2}{l}{\textit{Handcrafted descriptor}:} \\
        [.25ex]
        ICP & \textit{n/a} & 11.00 & 5.86 & 10.91 & 6.32 \\
        RANSAC & FPFH~\cite{rusu2009fpfh} & 0.87 & 1.13 & 1.35 & 3.55 \\
        TEASER++~\cite{yang2021teaser} & FPFH~\cite{rusu2009fpfh} & 0.04 & 0.10 & 0.54 & 3.24 \\
        PointDSC~\cite{bai2021pointdsc} & FPFH~\cite{rusu2009fpfh} & 0.06 & 0.15 & 2.91 & 9.70 \\
        \grayrule
        \multicolumn{4}{l}{\textit{Descriptors trained on 3DMatch~\cite{zeng20173dmatch}}:} \\
        [.25ex]
        RANSAC & DIP~\cite{poiesi2021dip} & 0.16 & 0.22 & 0.80 & 2.25 \\
        RANSAC & GeDi~\cite{poiesi2023gedi} & 0.50 & 0.73 & 1.71 & 4.25 \\
        \grayrule
        \multicolumn{4}{l}{\textit{Descriptors trained on KITTI~\cite{geiger2012kitti}}:} \\
        [.25ex]
        RANSAC & FCGF~\cite{choy2019fcgf} & 0.10 & 0.15 & 0.45 & 1.36 \\
        PointDSC~\cite{bai2021pointdsc} & FCGF~\cite{choy2019fcgf} & 0.07 & 0.14 & 0.51 & 1.64 \\
        RANSAC & SpinNet~\cite{ao2021spinnet} & 0.09 & 0.15 & 0.46 & 1.28 \\
        RANSAC & GCL~\cite{Liu2023gcl} & 0.09 & 0.14 & 0.46 & 1.26 \\
        \grayrule
        \multicolumn{2}{l}{\textit{Our method}:} \\
        [.25ex]
        RANSAC & DINOv2~\cite{oquab2024dinov2} & -- & -- & 0.56 & 1.44 \\
        \bottomrule
    \end{tabular}
    \footnotesize
    We underline the effect of a domain change on various point descriptors used as baselines in our experiments. While the descriptors trained on the KITTI dataset~\cite{geiger2012kitti} perform well within the same domain, they suffer from degradation when tested on the NCLT dataset~\cite{bianco2016nclt}. Note that PointDSC~\cite{bai2021pointdsc} is also trained on KITTI. Due to the lack of surround-view images, we do not employ our method on KITTI. In contrast to the primary use case of this work, previous studies mostly considered scan-to-scan registration.
\end{threeparttable}
\vspace*{-.4cm}
\end{table}

A core advantage of employing DINOv2~\cite{oquab2024dinov2} is exploiting its strong generalization capability across various semantic domains and decoupling the point descriptors from the density of a point cloud, e.g., RGB-D data versus LiDAR scans. Many learning-based descriptors suffer from such domain changes requiring in-domain training data and hindering their general applicability. To underline this effect and to establish the groundwork for the main experiment, we first report the performance of the baselines for the more simple scan-to-scan registration task. This task is mainly considered by previous studies in the context of LiDAR odometry. Here, the two point clouds are highly similar in their geometry and a strong initial guess is available. We simulate the initial guess by perturbing the ground truth transform with noise sampled as:
\begin{equation}
\begin{aligned}
    \mathbf{t}_x, \mathbf{t}_y &\sim \mathcal{N}(0, 10) \, , \mathbf{t}_y \sim \mathcal{N}(0, 1) \, , \\
    \mathbf{R}_x, \mathbf{R}_y &\sim \mathcal{N}(0, 2) \, , \mathbf{R}_z \sim \mathcal{N}(0, 10) \, ,
\end{aligned}
\end{equation}
with $\mathbf{t}$ and $\mathbf{R}$ referring to the translational and rotational components measured in meters and degrees, respectively.
In \cref{tab:sanity-exp}, we report the average RTE and RRE on samples from the KITTI and NCLT datasets. For KITTI, we sample 125 pairs of two consecutive scans from sequence 08, which is not in the descriptors' training set~\cite{ao2021spinnet, choy2019fcgf}. For NCLT, we use the scan of the map that is closest to the incoming LiDAR scan. Note that we do not employ our method on KITTI due to the lack of surround-view images. The key insights from \cref{tab:sanity-exp} are as follows: (1) On KITTI, the error of the in-domain trained descriptors is substantially smaller than of the ones trained on 3DMatch; (2) We observe poor performance when the training and testing domains differ, i.e., from 3DMatch to KITTI or NCLT, and from KITTI to NCLT; (3) Most descriptors achieve a decent accuracy on NCLT for the scan-to-scan registration task.
\looseness=-1

\begin{figure}
    \centering
    \includegraphics[width=1\linewidth]{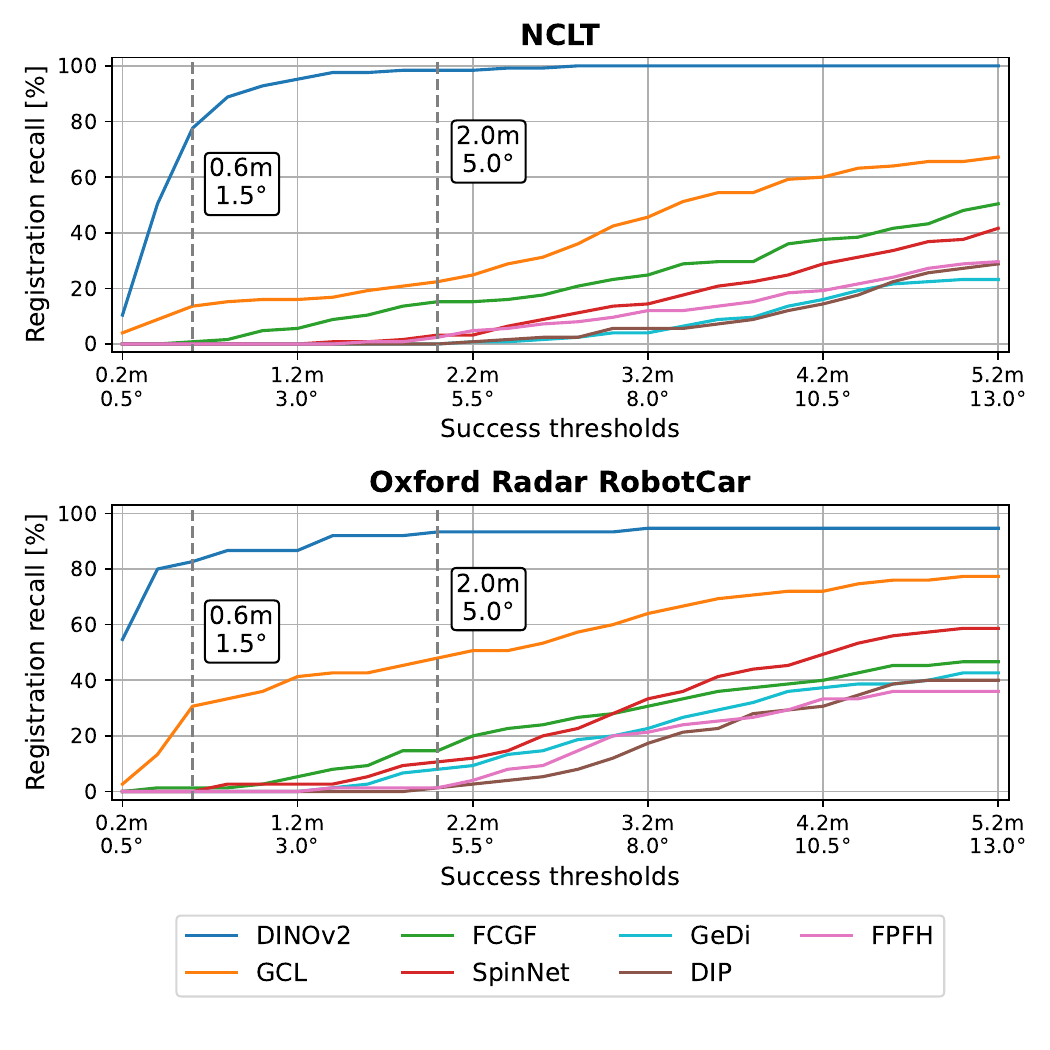}
    \vspace*{-1.0cm}
    \caption{We visualize the registration recall (RR) for a range of success thresholds obtained by linear inter/extrapolation of the thresholds used by GCL~\cite{Liu2023gcl} (\textit{left dashed line}) and SpinNet~\cite{ao2021spinnet} (\textit{right dashed line}). To perform scan-to-map registration, we couple RANSAC with the specified point descriptors.}
    \label{fig:registration-recall}
    \vspace*{-.5cm}
\end{figure}

In the main experiment, we support our claim that our proposed DINOv2-based point descriptors can be coupled with traditional registration schemes while outperforming previous baselines. We now consider scan-to-map registration using the extracted scenes as described in \cref{ssec:experimental-setup}. 
We report results for the metrics defined in \cref{ssec:metrics} for both the NCLT dataset and the Oxford Radar RobotCar dataset in \cref{tab:scan-to-map-exp}. The most important observation is that our proposed method is the only one that achieves consistently low registration errors. For the ICP-based refinement, our method substantially outperforms the best baseline (GCL~\cite{Liu2023gcl}) by $24.8$ (NCLT) and $17.3$ (RobotCar) percentage points, showing \SI{100}{\percent} recall on NCLT.
We hypothesize that the gap to \SI{100}{\percent} on the Oxford Radar RobotCar is mainly caused by wrong point-to-pixel projections, e.g., due to erroneous extrinsic calibration, jerky ego-motion, and differences in the sensors' viewpoints. An indicator for this hypothesis is the observation that some points belonging to buildings are assigned tree-like descriptors if there is a tree in front of the wall.
For completeness, we also report results when replacing RANSAC with TEASER++~\cite{yang2021teaser}, achieving higher registration recalls than all baseline methods.
A further key insight is the large standard deviation of the errors of the baseline methods, whereas the results of our method rarely fluctuate. Finally, we note that the baselines yield almost no global registration meeting the success thresholds, resulting in \SI{0.00}{\percent} registration recall. To further investigate this observation and to incorporate more relaxed thresholds as used in other studies, we recompute the registration recall (RR) for additional thresholds. In particular, we use linear inter/extrapolation of the thresholds based on GCL~\cite{Liu2023gcl} (\SI{0.6}{\meter}, \SI{1.5}{\degree}) and SpinNet~\cite{ao2021spinnet} (\SI{2.0}{\meter}, \SI{5.0}{\degree}). We visualize the registration recalls in \cref{fig:registration-recall} for all descriptors coupled with RANSAC. The recall of our DINOv2-based descriptors yields a high recall even for strict thresholds, eventually converging towards the recall after ICP refinement. The recall of the other baselines slowly increases for large success thresholds, failing to achieve accurate map-based localization.

\begin{figure*}
    \centering
    \includegraphics[width=1.\linewidth]{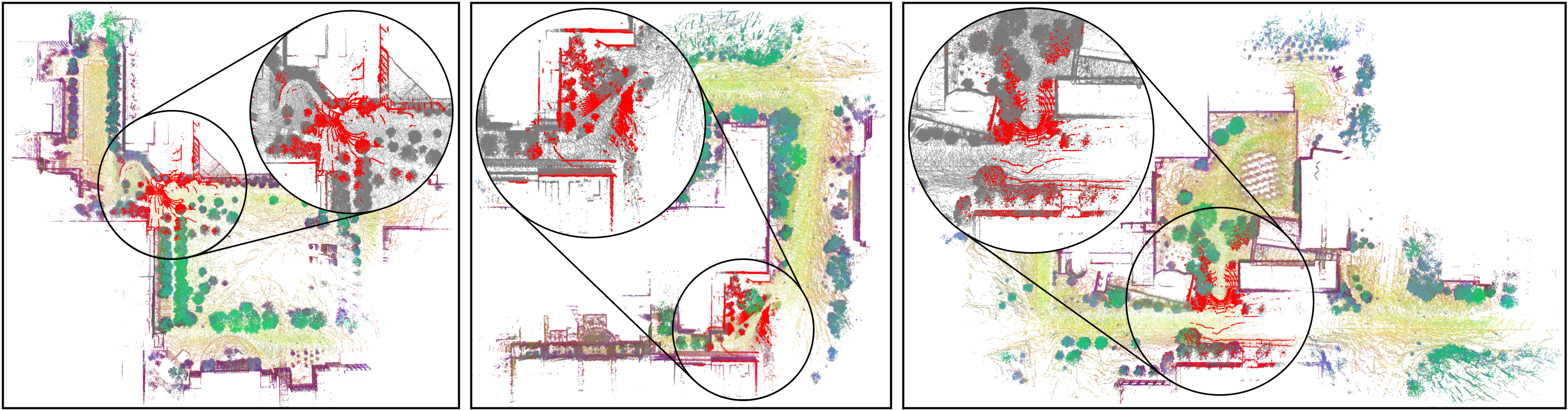}
    \vspace*{-.6cm}
    \caption{Qualitative results on the NCLT dataset~\cite{bianco2016nclt}. We colorize the 3D map by taking the first three components of performing principal component analysis~(PCA) on the point descriptors of the map. The successfully registered LiDAR scan is shown in red.}
    \label{fig:results-nclt}
\end{figure*}

\begin{figure*}
    \centering
    \includegraphics[width=1.\linewidth]{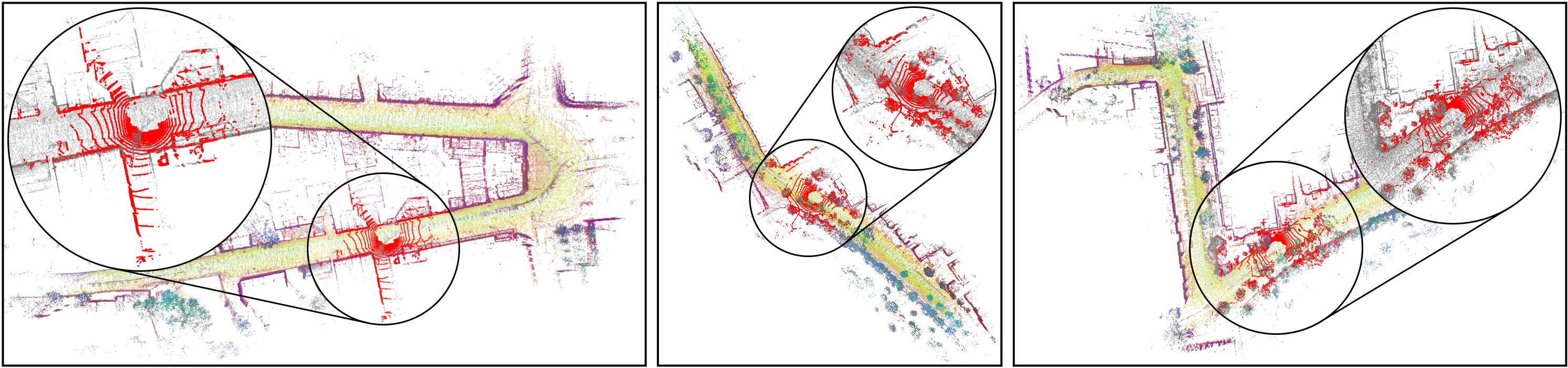}
    \vspace*{-.6cm}
    \caption{Qualitative results on the Oxford Radar RobotCar dataset~\cite{barnes2020robotcarradar}. We colorize the 3D map by taking the first three components of performing principal component analysis~(PCA) on the point descriptors of the map. The successfully registered LiDAR scan is shown in red.}
    \label{fig:results-robotcar}
\end{figure*}

We conclude this experiment by visualizing successful registrations in \cref{fig:results-nclt,fig:results-robotcar}. Note that the colors of the 3D map are obtained via principal component analysis of the initial NCLT scene. Following previous work~\cite{oquab2024dinov2}, we adopt the first three components as color channels in the RBG space. To identify the registered LiDAR scan within the 3D map, we do not employ this colorization to the scan but show it in red.

%%%%%%%%%%%%%%%%%%%%%%%%%%%%%%%%%%%%%%%%%%%%%%%%%%%%%%%%%%%%%%%%%%%%%%%%%%%%%%%%

\begin{figure}
    \centering
    \includegraphics[width=\linewidth]{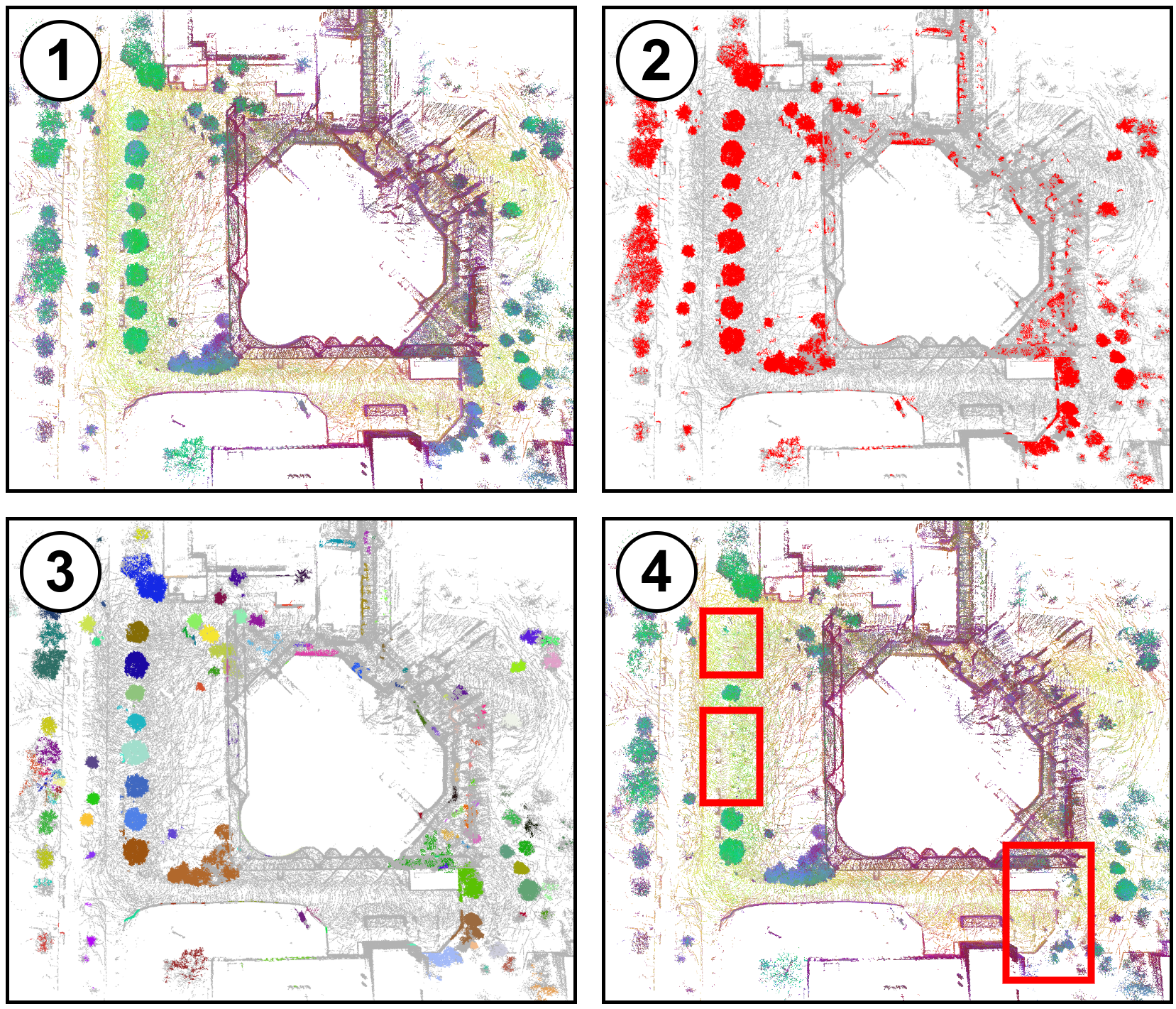}
    \vspace*{-.6cm}
    \caption{To show the robustness of our proposed approach, we remove semantic entities from the 3D map. (1) The original map. (2) We identify tree-like points colored in red using the DINOv2-based descriptors. (3) We assign these points to separate clusters shown in different colors. (4) We randomly remove some clusters from the 3D map, highlighted by the red boxes. \looseness=-1}
    \label{fig:outdated-map}
    \vspace*{-.3cm}
\end{figure}

\subsection{Robustness to Environmental Changes}

This experiment is designed to support our third claim, i.e., the DINOv2-based descriptors are robust to temporal environmental changes resulting in outdated map data.
While the previous experiment provides some insight due to the seasonal variations, e.g., snow and foliage in the NCLT dataset~\cite{bianco2016nclt}, we attempt to further amplify long-term changes. On the NCLT dataset, we carefully remove distinct objects from the 3D map following the steps visualized in \cref{fig:outdated-map}. (1) Given an input map, (2) we initially query all tree-like points on the map by considering the similarity of the point descriptors with the descriptor of a point that is identified as part of a tree by a human supervisor. For simplicity, we adopt the Euclidean distance in the sRGB space after converting the point descriptors to the same PCA space as used in the visualization throughout this manuscript. If the distance is less than 50, we assume the point to be part of a tree. (3) Afterward, we use HDBSCAN~\cite{mcinnes2017hdbscan} to cluster the tree-like points using their 3D coordinates. (4) Finally, for each cluster, we decide with a given probability whether to remove it from the map. Since the urban scenes of the Oxford Radar RobotCar dataset~\cite{barnes2020robotcarradar} contain fewer trees, we invert the idea. In particular, we insert up to 100 additional trees in the scenes that we previously extracted from an NCLT scene.

\begin{figure}
    \centering
    \includegraphics[width=\linewidth]{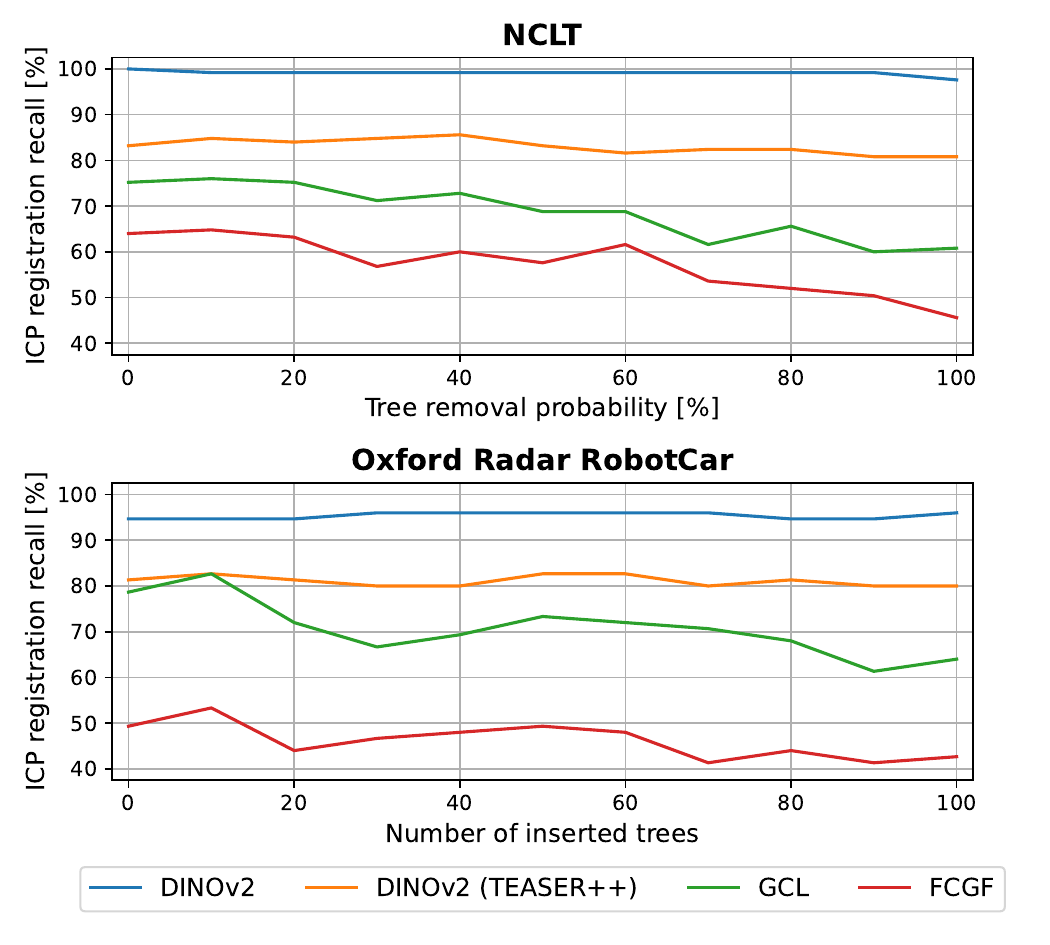}
    \vspace*{-0.8cm}
    \caption{We visualize the registration recall after ICP refinement (ICP-RR) for the removal and insertion of trees into the 3D map. Unlike the baseline methods, our proposed DINOv2-based point descriptor results in stable registration underlining its robustness.}
    \label{fig:robustness-recall}
    \vspace*{-.3cm}
\end{figure}

In \cref{fig:robustness-recall}, we report the registration recall after ICP refinement (ICP-RR) for both experiments using RANSAC and TEASER++~\cite{yang2021teaser} with our proposed point descriptors. We further compute the scores for the two best-performing baselines (see \cref{tab:scan-to-map-exp}), i.e., GCL~\cite{Liu2023gcl} and FCGF~\cite{choy2019fcgf}. While the baseline descriptors suffer from some degradation due to increasing changes in the reference map, our DINOv2-based point descriptor results in stable registration throughout the experiments, supporting our claim of its robustness.

%%%%%%%%%%%%%%%%%%%%%%%%%%%%%%%%%%%%%%%%%%%%%%%%%%%%%%%%%%%%%%%%%%%%%%%%%%%%%%%%

\section{Discussion of the Limitations}

In this section, we outline some limitations of our approach. As discussed in \cref{sec:intro}, LiDAR-camera sensor configurations are a common setup on mobile robots. We further show in \cref{ssec:exp-scan-to-map} that the additional vision modality enables our method to substantially outperform LiDAR-only baselines. Nonetheless, fusing these modalities introduces new challenges such as accurate extrinsic calibration and time synchronization. A primary source of error arises from incorrect projection of DINOv2 features from image pixels to the wrong LiDAR points, which can lead to misalignments, such as a \textit{tree} pixel being projected onto a building in the point cloud. Eventually, this results in an incorrect semantic description of a point. Other factors contributing to such errors include the presence of moving objects and varying fields of view of the sensors. Furthermore, limitations inherent to DINOv2, such as dependence on adequate illumination, may also affect performance. Lastly, we anticipate reduced performance when dealing with out-of-domain data, e.g., deployment in extraterrestrial environments like Mars. In most relevant scenarios though, this limitation is mitigated by the robust generalization capabilities of DINOv2.

\section{Conclusion}

In this study, we demonstrated that features extracted from DINOv2 using surround-view image data can be reliably utilized to establish correspondences between point clouds. Through extensive experimentation, we showed that integrating these DINOv2-based point descriptors with traditional registration methods, such as RANSAC or ICP, substantially enhances scan-to-map registration, outperforming various handcrafted and learning-based baselines. We also verified the robustness of the descriptor against seasonal variations and long-term environmental changes.
Future research will explore the potential application of visual foundation models directly to point cloud projections, eliminating the need for surround-view cameras and addressing challenges related to insufficient illumination and inaccuracies in point-to-pixel projection. Additionally, another direction for further research is explicitly combining the semantic richness of DINOv2 features with geometric features.

\section*{Acknowledgment}

This work was partially supported by a fellowship of the German Academic Exchange Service (DAAD).
Niclas Vödisch acknowledges travel support from the European Union’s Horizon 2020 research and innovation program under ELISE grant agreement No.~951847 and from the ELSA Mobility Program within the project European Lighthouse On Safe And Secure AI (ELSA) under the grant agreement No.~101070617.

\bibliographystyle{plainnat}
\bibliography{references}

\end{document}